\documentclass{llncs}
\usepackage{graphicx} %
\usepackage{xspace} %
\usepackage{color} %
\usepackage{amsmath}
\usepackage{amsfonts}
\usepackage{amssymb}
\usepackage{paralist}
\usepackage[labelfont={sc}]{subfig}

\usepackage{pdflscape}
\usepackage{setspace}
\usepackage{pifont}

\usepackage[us]{datetime}
\usepackage{fancyhdr}
\usepackage{wrapfig}
\usepackage{sidecap}
\usepackage{fancybox}
\usepackage{afterpage}

\newsavebox\FrameBox

\newcommand{\rf}{{receptive field}\xspace}%
\newcommand{\rfs}{{receptive fields}\xspace}%
\newcommand{\Rfs}{{Receptive fields}\xspace}%

\newcommand{\ie}{i.e.,\xspace} %
\newcommand{\pd}{{\partial}} %
\newcommand{\fig}{Fig\xspace} %
\newcommand{\figs}{Figs\xspace} %
\newcommand{\am}{{apparent motion}\xspace} %
\newcommand{\Real}{\mathbb{R}} %
\newcommand{\eq}{Eq.\@\xspace} %
\newcommand{\eqs}{Eqs.\@\xspace} %
{\begin{Sbox}\begin{minipage}}%
{\end{minipage}\end{Sbox}\fbox{\TheSbox}}

\everymath{\textstyle}

\begin{document}

\title{Optimal measurement of visual motion \\ across spatial and temporal scales}

\author{Sergei Gepshtein\inst {1} \and Ivan Tyukin \inst{2,3}}

\institute{
Systems Neurobiology Laboratories, Salk Institute for Biological Studies\\
10010 North Torrey Pines Road, La Jolla, CA 92037, USA\\
\texttt{sergei@salk.edu}
\vspace{.1in}
\and Department of Mathematics, University of Leicester\\
University Road, Leicester LE1 7RH, UK\\
\texttt{i.tyukin@le.ac.uk}
\vspace{.1in}
\and Department of Automation and Control Processes\\
Saint-Petersburg State Electrotechnical University\\
5 Professora Popova Str., 197376, Saint-Petersburg, Russia
}

\maketitle

\begin{center}
{\small April 26, 2014}
\end{center}

\begin{abstract}
Sensory systems use limited resources to mediate the perception of a great variety of  objects and events. Here a normative framework is presented for exploring how the problem of efficient allocation of resources can  be solved in visual perception. Starting with a basic property of every measurement, captured by Gabor's uncertainty relation about the location and frequency content of signals, prescriptions are developed  for optimal allocation of sensors for reliable perception of visual motion. This study reveals that a large-scale characteristic of human vision (the spatiotemporal contrast sensitivity function) is similar to the optimal prescription, and it suggests that some previously puzzling phenomena of visual sensitivity, adaptation, and perceptual organization have simple principled explanations.
\end{abstract}

\section{Introduction}

Biological sensory systems collect information from a vast range of spatial and temporal scales. For example, human vision can discern modulations of luminance that span nearly seven octaves of spatial and temporal frequencies, while many properties of optical stimulation (such as the speed and direction of motion) are analyzed within every step of the scale.

The large amount of information is encoded and transformed for the sake of specific visual tasks using limited resources. In biological systems, it is a large but finite number of neural cells. The cells are specialized: sensitive to a small subset of optical signals, presenting sensory systems with the problem of allocation of limited resources. This chapter is concerned with how this problem is solved by biological vision. How are the specialized cells distributed across the great number of potential optical signals in the environments that are diverse and variable?

The extensive history of vision science suggests that any attempt of vision theory should begin with an analysis of the tasks performed by visual systems.  Following Aristotle, one may begin with the definition of vision as ``knowing hat is where by looking" \cite{Marr1982}. The following argument concerns the basic visual tasks captured by this definition.

The ``what" and ``where" of visual perception are associated with two characteristics of optical signals: their frequency content and  locations, in space and time. The last statement implicates at least five dimensions of optical signals (which will become clear in a moment).

The basic visual tasks are bound by first principles of measurement. To see that, consider a measurement device (a ``sensor" or ``cell") that integrates its inputs over some spatiotemporal interval. An individual device of an arbitrary size will be more suited for measuring the location \emph{or} the frequency content of the signal, reflected in the uncertainties of measurement. The uncertainties associated with the location and the frequency content are related by a simple law formalized by Gabor \cite{Gabor1946}, who showed that the two uncertainties trade off across scales. As the scale changes, one uncertainty rises and the other falls.

Assuming that the visual systems in question are interested in both the locations and frequency content of optical signals (``stimuli"), the tradeoff of uncertainties will attain a desired (``optimal") balance of uncertainties at some intermediate scale. The notion of the optimal tradeoff of uncertainty has received considerable attention in studies of biological vision. This is because the ``receptive fields" of single neural cells early in the visual pathways appear to approximate one or another form of the optimal tradeoff  \cite{Marcelja1980,MacKay1981,Daugman1985,Glezer1986,Field1987,JonesPalmer1987,Simoncelli_Olshausen2001,Saremi_etal2013}.

Here the tradeoff of uncertainties is formulated in a manner that is helpful for investigating its consequences outside of the optimum: across many scales, and for cell populations rather than for single cells. Then the question is posed of how the scales of multiple sensory cells should be selected for simultaneously minimizing the uncertainty of measurement for all the cells, on several stimulus dimensions.

The present focus is on how visual motion can be estimated at the lowest overall uncertainty of measurement across the entire range of useful sensor sizes (in artificial systems) or the entire range of receptive fields (in biological systems).  In other words, the following is an attempt to develop an economic \emph{normative theory} of motion-sensitive systems. Norms are derived for efficient design of such systems, and then the norms are compared with facts of biological vision.

This approach from first principles of measurement and parsimony helps to understand the forces that shape the characteristics of biological vision, but which had appeared intractable or controversial using previous methods. These characteristics include the spatiotemporal contrast sensitivity function, adaptive transformations of this function caused by stimulus change, and also some characteristics of the higher-level perceptual processes, such as perceptual organization.

\begin{figure}[p]
\centering
\includegraphics[width=0.47\textwidth]{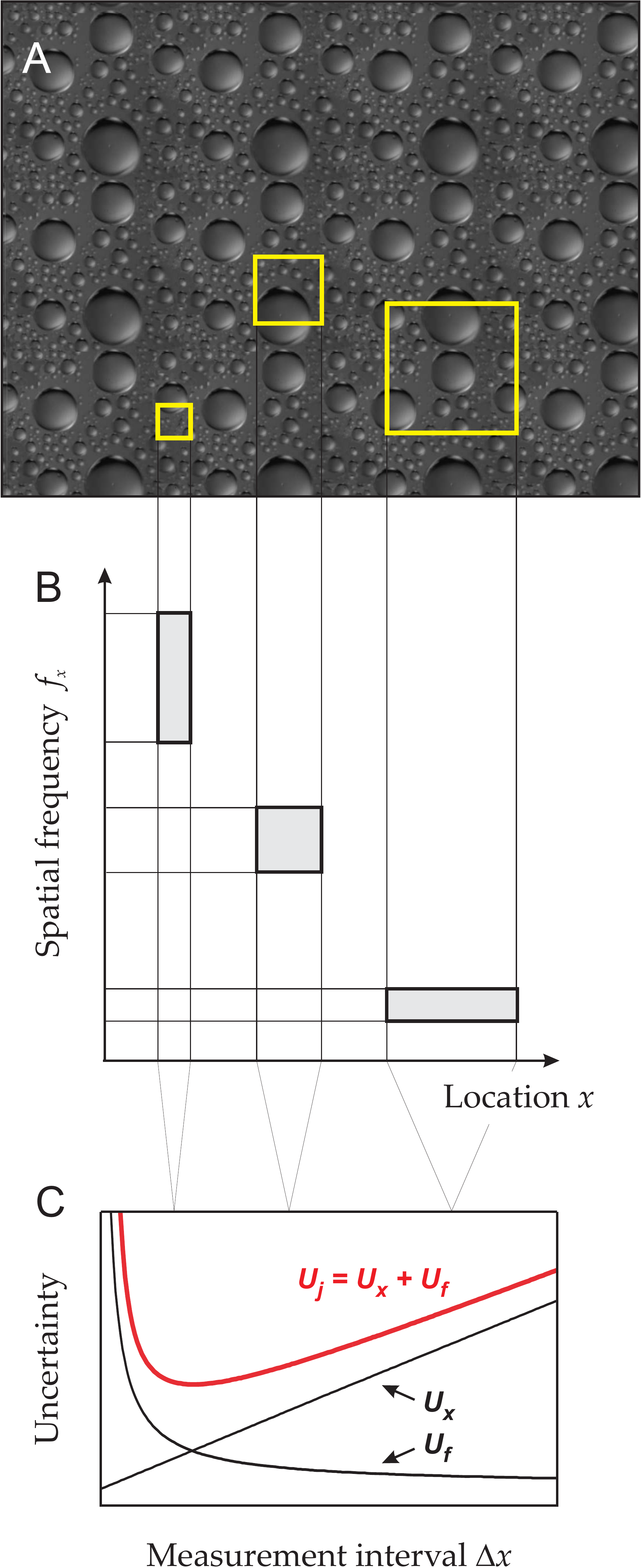}
\vspace{.2in}
\caption{\setstretch{1.0} Components of measurement uncertainty. (A)~The image is sampled by three sensors of different sizes. (B)~The three sensors are associated with Gabor's \emph{logons}: three rectangles that have the same areas but different shapes, according to the limiting condition of the uncertainty relation in \eq~\ref{eq:uncert_eq}. (C)~Functions $U_x$ and $U_f$ represent the uncertainties about the location and content of the measured signal (the horizontal and vertical extents of the logons in~B, respectively), and function $U_j$ represents the joint uncertainty about signal location and content.}
\label{fig:logon1}
\end{figure}

\section{Gabor's uncertainty relation in one dimension}

\noindent
The outcomes of measuring the location and the frequency content of any signal by a single sensory device are not independent of one another. The measurement of location assigns the signal to interval $\Delta x$ on some dimension of interest $x$. The smaller the interval the lower the uncertainty about signal location. The uncertainty is often described in terms of the precision of measurements, quantified by the dispersion of the measurement interval or, even simpler, by the size of the interval, $\Delta x$. The smaller the interval, the lower the uncertainty about location, and the higher the precision of measurement.

The measurement of frequency content evaluates how the signal varies over $x$, \ie the measurement is best described  on the dimension of frequency of signal variation, $f_x$.  That is, the measurement of frequency content is equivalent to localizing the signal on $f_x$: assigning the signal to some interval $\Delta f_x$. Again, the smaller the interval, the lower the uncertainty of measurement and the higher the precision.\footnote{For brevity, here ``frequency content" will sometimes be shortened to ``content."}

The product of uncertainties about the location and frequency content of the signal is bounded ``from below" \cite{Gabor1946,Gabor1952,Resnikoff1989,MacLennan91}. The product cannot be smaller than some positive constant $C_x$:
\begin{equation}
U_x U_f  \ge  C_x,
\label{eq:uncert_U}
\end{equation}
where $U_x$ and $U_f$ are the uncertainties about the location and frequency content of the signal, respectively, measured on the intervals $\Delta x$ and $\Delta f_x$.

\eq~\ref{eq:uncert_U} means that any measurement has a limit at $U_x U_f = C_x$.  At the limit, decreasing  one uncertainty is accompanied by increasing the other. For simplicity, let us quantify the  measurement uncertainty by the size of the measurement interval.  Gabor's uncertainty relation may therefore be written as
\begin{equation}
\Delta x \Delta f_x \ge  C_x,
\end{equation}
and its limiting condition as
\begin{equation}
\Delta x \Delta f_x =  C_x.
\label{eq:uncert_eq}
\end{equation}

\subsection{Single sensors}

Let us consider how the uncertainty relation constrains the measurements by a single measuring device: a ``sensor."  \fig~\ref{fig:logon1} illustrates three spatial sensors of different sizes. In \fig~\ref{fig:logon1}A, the measurement intervals of the sensors are defined on two spatial dimensions. For simplicity, let us consider just one spatial dimension, $x$, so the interval of measurement (``sensor size") is $\Delta x$, as in \fig~\ref{fig:logon1}B--C.

The limiting effect of the uncertainty relation for such sensors has a convenient graphic representation called ``information
diagram" (\fig~\ref{fig:logon1}B). Let the two multiplicative terms of \eq~\ref{eq:uncert_eq} be represented by the two sides of a rectangle in coordinate plane ($x$, $f_x$). Then $C_x$ is the rectangle area. Such rectangles are called ``information cells" or ``logons." Three logons, of different shapes but of the same area $C_x$, are shown in \fig~\ref{fig:logon1}B, representing the three sensors:
\begin{itemize}
\item
The logon of the smallest sensor (smallest $\Delta x$, left) is thin and tall, indicating that the sensor has a high precision on $x$
and a low precision on $f_x$.
\vspace{.1in}
\item
The logon of the largest sensor (right) is thick and short, indicating a low precision on $x$ and a high precision on $f_x$.
\vspace{.1in}
\item
The above sensors are specialized for measuring either the location or frequency content of  signals. The medium-size sensor (middle) offers a compromise: its uncertainties are not as low as the lowest uncertainties (but not as high as the highest uncertainties) of the specialized sensors. In this respect, the medium-size sensor trades one kind of uncertainty for another.
\end{itemize}
The medium-size sensors are most useful for jointly measuring the locations and frequency content of signals.

So far, the ranking of sensors has been formalize using an additive model of uncertainty (\fig~\ref{fig:logon1}C). The motivation for such an additive model is presented in Appendix~1. This approach is motivated by the assumption that visual systems have no access to complete prior information about the statistics of measured signals (such as the joint probability density functions for the spatial and temporal locations of stimuli and their frequency content). The assumption is, instead, that the systems can reliably estimate only the means and variances of the measured quantities.

Accordingly, the overall uncertainty in \fig~\ref{fig:logon1}C has the following components. The increasing function represents the uncertainty about signal location: $U_x = \Delta x$.  The decreasing function represents the uncertainty about signal content: $U_f = \Delta f_x = C_x / \Delta x\;$ (from \eq~\ref{eq:uncert_eq}).
The joint uncertainty of measuring signal location \emph{and} content is represented by the non-monotonic function $U_j$:
\begin{equation}
U_j =  \lambda_x U_x  +  \lambda_f U_f = \lambda_x \Delta x  +  \lambda_f \frac{1}{\Delta x},
\label{eq:u_j}
\end{equation}
where $\lambda_x $ and $\lambda_f$ are positive coefficients reflecting how important the components of uncertainty are relative to one another.

The additive model of \eq~\ref{eq:u_j} implements a worst-case estimate of the overall uncertainty (as it is explained in section~\emph{The Minimax Principle} just below). The additive components are weighted, while the weights are playing several roles. They bring the components of uncertainty to the same units, allowing for different magnitude of $C_x$,\footnote{Different criteria of measurement and sensor shapes correspond to different magnitudes of $C_x$.} and representing the fact that the relative importance of the components depends on the task at hand.

The joint uncertainty function ($U_j$ in \fig~\ref{fig:logon1}C) has its minimum at an intermediate value of $\Delta x$.  This is a \emph{point of equilibrium} of uncertainties, in that a sensor of this size implements a perfect balance of uncertainties about the location and frequency content of the signal \cite{GepshteinTyukin2006}. If measurements are made in the interest of high precision, and if the location and the frequency content of the signal are equally important, then a sensor of this size is the best choice for jointly measuring the location and the frequency content of the signal.

\paragraph{The Minimax Principle.}  What is the best way to allocated resources in order to reduce the chance of gross errors of measurement. One approach to solving this problem is using the minimax strategy devised in game theory for modeling choice behavior \cite{vonNeumann1928,LuceRaiffa57}. Generally, the minimax strategy is used for estimating the maximal expected loss for every choice and then pursuing the choices, for which the expected maximal loss is minimal. In the present case, the choice is between the sensors that deliver information with variable uncertainty.

In the following, the minimax strategy is implemented by assuming the maximal (worst-case) uncertainty of measurement on the sensors that span the entire range of the useful spatial and temporal scales. This strategy  is used in two ways. First, the consequences of Gabor's uncertainty relation are investigated by assuming that the uncertainty of measurement is as high as possible (\ie using the limiting case of uncertainty relation; \eq~\ref{eq:uncert_eq}). Second, the outcomes of measurement on different sensors are anticipated by \emph{adding} their component uncertainties, \ie using the joint uncertainty function of \eq~\ref{eq:u_j}. (The choice of the additive model is explained in Appendix~1.) It is assumed that sensor preferences are ranked according to the expected maximal uncertainty: the lower the uncertainty, the higher the preference.

\begin{figure}[p]
\centering
\includegraphics[width=0.55\textwidth]{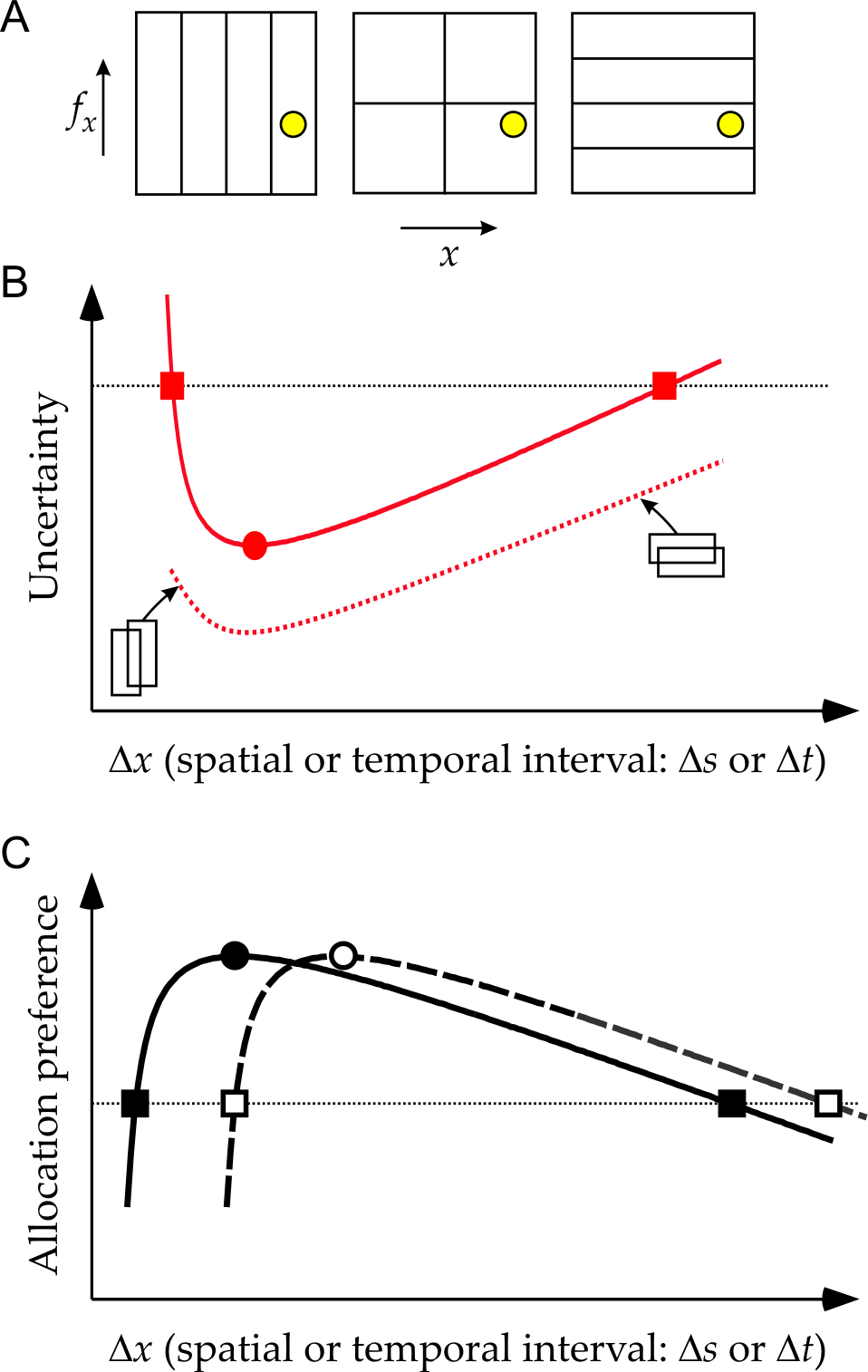}
\vspace{.1in}
\caption{\setstretch{1.0} Allocation of multiple sensors. (A)~Information diagrams for a population of four sensors, using sensors of the same size within each population, and of different sizes across the populations. (B)~Uncertainty functions. The red curve is the joint uncertainty function introduced in \fig~\ref{fig:logon1}, with the markers indicating special conditions of measurement: the lowest joint uncertainty (the circle) and the equivalent joint uncertainty (the squares), anticipating the \emph{optimal sets} and the \emph{equivalence classes} of measurement in the higher-dimensional systems illustrated in \figs~\ref{fig:many_logons3d}--\ref{fig:eq_contours}. (C)~Preference functions. The solid curve is a function of allocation preference (here reciprocal to the uncertainty function in B): an optimal distribution of sensors, expected to shift (dashed curve) in response to change in stimulus usefulness.}
\label{fig:many_logons1d}
\end{figure}

\subsection{Sensor populations}

Real sensory systems have at their disposal large but limited numbers of sensors. Since every sensor is useful for measuring only some aspects of the stimulus, sensory systems must solve an economic problem: they must distribute their sensors in the interest of perception of many different stimuli.  Let us consider this problem using some simple arrangements of sensors.

First, consider a population of identical sensors in which the measurement intervals do not overlap. \fig~\ref{fig:many_logons1d}A contains three examples of such sensors, using the information diagram introduced in \fig~\ref{fig:logon1}. Each of the three diagrams in \fig~\ref{fig:many_logons1d}A portrays four sensors, identical to one another except they are tuned to different intervals on~$x$ (which can be space or time).  Each panel also contains a representation of a narrow-band signal: the yellow circle, the same across the three panels of \fig~\ref{fig:many_logons1d}A. The different arrangements of sensors imply different \emph{resolutions} of the system for measuring the location and frequency content of the stimulus.
\begin{itemize}
\item
The population of small sensors (small $\Delta x$ on the left of \fig~\ref{fig:many_logons1d}A)  is most suitable for measuring signal  location: the test signal is assigned to the rightmost quarter on
the range of interest in~$x$. In contrast, measurement of frequency content is poor: signals presented anywhere within the vertical extent of the sensor (\ie within the large interval on $f_x$) will all give the same response. This system has a good location resolution and poor frequency resolution.
\vspace{.1in}
\item The population of large sensors (large $\Delta x$ on the right of \fig~\ref{fig:many_logons1d}A) is most suitable  for measuring frequency content.  The test signal is assigned to a small interval on~$f_x$. Measurement of location is poor. This system has a good frequency resolution and poor location resolution.
\vspace{.1in}
\item The population of medium-size sensors can obtain useful information about both locations and frequency content of signals. It has a better frequency resolution than the population of small sensors, and a better location resolution than the population of large sensors.
\end{itemize}

Consequences of the different sensor sizes are summarized by the joint uncertainty function in \fig~\ref{fig:many_logons1d}B. (For non-overlapping sensors, the function has the same shape as in \fig~\ref{fig:logon1}C). The figure makes it clear that the sensors or sensor populations with very different properties can be equivalent in terms of their joint uncertainty. For example, the two filled squares in \fig~\ref{fig:many_logons1d}B mark the uncertainties of two different sensor populations: one contains only small sensors and the other contains only large sensors.

The populations of sensors in which the measurement intervals overlap are more versatile than the populations of non-overlapping sensors. For example, the sensors with large overlapping  intervals can be used to emulate measurements by the sensors with smaller intervals (Appendix~2), reducing the uncertainty of stimulus localization. Similarly, groups of the overlapping sensors with small measurement intervals can emulate the measurements by sensors with larger intervals, reducing the uncertainty of identification.  Overall, a population of the overlapping sensors can afford lower uncertainties across the entire range of measurement intervals, represented in \fig~\ref{fig:many_logons1d}B  by the dotted curve: a lower-envelope uncertainty function. Still, the new uncertainty function has the same shape as the previous function (represented by the solid line) because of the limited total number of the sensors.

\subsection{Cooperative measurement}

To illustrate the benefits of measurement using multiple sensors, suppose that the stimulation was uniform and one could vary the number of sensors in the population at will, starting with a system that has only a few sensors, toward a system that has an unlimited number of sensors.
\begin{itemize}
\item A system equipped with very limited resources, and seeking to measure both the location and the frequency content of signals, will have to be unmitigatedly frugal.  It will use only the sensors of medium size, because only such sensors offer useful (if limited) information about both properties of signals.

\vspace{.1in} \item
A system enjoying unlimited resources, will be able to afford many specialized sensors, or groups of such sensors (represented by the information diagrams in \fig~\ref{fig:many_logons1d}A).

\vspace{.1in} \item
A moderately wealthy system: a realistic middle ground between the extremes outlined above, will be able to escape the straits of Gabor's uncertainty relation using different specialized sensors and thus measuring the location and content of signals with a high precision.
\end{itemize}
As one considers systems with different numbers of sensors, from small to large, one expects to find an increasing ability of the system to afford the large and small measurement intervals.  As the number of sensors increases, the allocation of sensors will expand in two directions, up and down on the dimension of sensor scale: from using only the medium-size sensors in the poor system, to using also the small and large sensors in the wealthier systems. This allocation policy is illustrated in \fig~\ref{fig:many_logons1d}C. The \emph{preference function} in \fig~\ref{fig:many_logons1d}C indicates that, as the more useful sensors are expected to grow in number, the distribution of sensors will form a smooth function across the scales. As mentioned, the sensitivity of the system is expected to follow a function monotonically related to the preference function.

Increasing the number of sensors selective to the same stimulus condition is expected to improve sensory performance, manifested in lower sensory thresholds. One reason for such improvement in biological sensory systems is the fact that integrating information across multiple sensors will help to reduce the detrimental effect of the noisy fluctuations of neural activity, in particular when the noises are uncorrelated.

The preference function in \fig~\ref{fig:many_logons1d}C is exceedingly simple: it merely mirrors the joint uncertainty function of \fig~\ref{fig:many_logons1d}B. This example helps to illustrate some special conditions of the uncertainty of measurement and to anticipate their consequences for sensory performance. First, the minimum of uncertainty corresponds to the maximum of allocation preference, where the highest sensitivity is expected. Second, equal uncertainties correspond to equal allocation preferences, where equal sensitivities are expected. Allocation policies are considered again in Sections~\ref{sec:optimal_conditions}--\ref{sec:allocation}, where the relationship is studied between a normative prescription for resource allocation and a characteristic of performance in biological vision.

\section{Gabor's uncertainty in space-time}

\subsection{Uncertainty  in two dimensions}
\begin{figure}[p]
\centering
\includegraphics[width=0.70\textwidth]{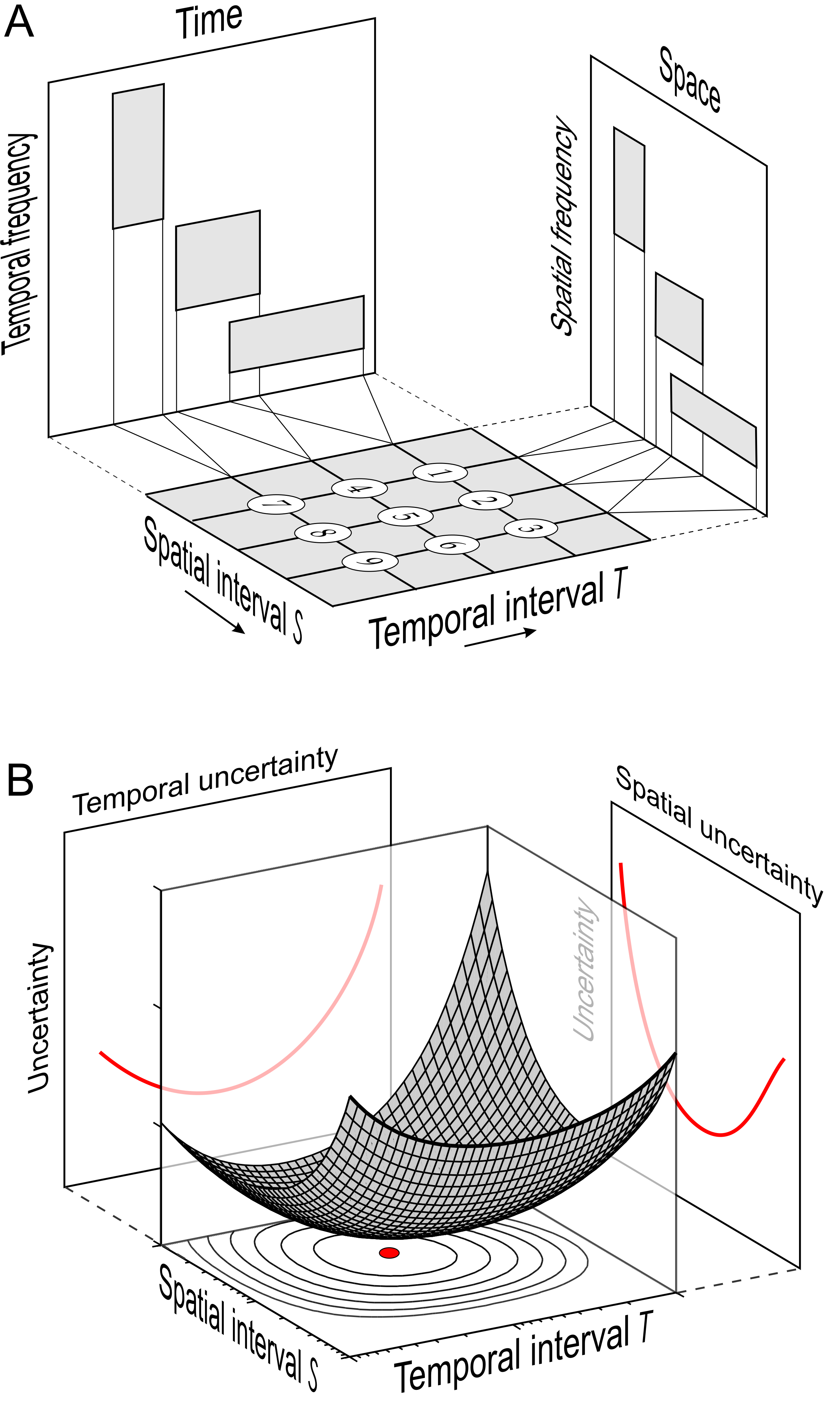}
\vspace{.1in}
\caption{\setstretch{1.0} Components of measurement uncertainty in space-time. (A)~Spatial and temporal information diagrams of spatiotemporal measurements. The numbered discs each represents a sensor of particular spatial and
temporal extent, $S=\Delta s$ and  $T=\Delta t$. The rectangles on side panels are the spatial and temporal logons associated with the sensors.
(B)~The surface represents the joint uncertainty about signal location and frequency content of signals across sensors of different spatial and temporal size. The contours in the bottom plane ($S$, $T$) are sets of
equivalent uncertainty (reproduced for further consideration in \fig~\ref{fig:eq_contours}). Panel~A is adopted from \cite{Gepshtein2010} and panel~B from \cite{GepshteinTyukinKubovy2007}.
}
\label{fig:many_logons3d}
\end{figure}

Now consider a more complex case where signals vary on two dimensions: space and time.  Here, the measurement uncertainty has four components, illustrated in \fig~\ref{fig:many_logons3d}A. The bottom of \fig~\ref{fig:many_logons3d}A is a graph of the spatial and temporal sensor sizes $(T, S) =
(\Delta t, \Delta s)$.  Every point in this graph corresponds to a ``condition of measurement" associated with the four properties of sensors.\footnote{Here the sensors are characterized by intervals following the standard notion that biological motion sensors are maximally activated when the stimulus travels some distance $\Delta s$ over some temporal interval $\Delta t$ \cite{WatsonAhumada1985}.} By Gabor's uncertainty relation, spatial and temporal intervals  $(\Delta t, \Delta s)$ are associated with, respectively, the spatial and temporal frequency intervals $(\Delta f_t, \Delta f_s)$.

The four-fold dependency is explained on the side panels of the figure using Gabor's logons, each associated with a sensor labeled by a numbered disc. For example,  in sensor~7 the spatial and temporal intervals are small, indicating a good precision of spatial and temporal localization (\ie~concerning ``where" and ``when" the stimuli occurs). But the spatial and temporal frequency intervals are large, indicating a low precision in measuring spatial and temporal frequency content (a low capacity to serve the ``what" task of stimulus identification). This pattern is reversed in sensor~3, where  the precision of localization is low but the precision of identification is high.

As in the previous example (\fig~\ref{fig:logon1}B--C), here the one-dimensional uncertainties are summarized using  joint uncertainty functions: the red curves on the side panels of \fig~\ref{fig:many_logons3d}B.
Each function has the form of \eq~\ref{eq:u_j}, applied separately to spatial: \[U_S =  \lambda_1 S +  \lambda_2 /S\] and temporal: \[ U_T =  \lambda_3 T +  \lambda_4 /T, \]
dimensions, where $S = \Delta s$ and $T = \Delta t$.  Next, spatial and temporal uncertainties are combine for every spatiotemporal condition: \[U_{ST} = U_T + U_S,\] to obtain a bivariate \emph{spatiotemporal uncertainty function}:
\begin{equation}
U_{ST} =  \lambda_1 S +  \frac{\lambda_2}{S} +  \lambda_3 T +  \frac{\lambda_4}{T}
\label{eq:u_st}
\end{equation}
represented in \fig~\ref{fig:many_logons3d}B by a surface.

The spatiotemporal uncertainty function in \fig~\ref{fig:many_logons3d}B has a unique minimum, of which the projection on graph $(T, S)$  is marked by the red dot: the point of perfect balance of the four components of measurement uncertainty.  Among the conditions of imperfect balance of uncertainties, consider the conditions of an equally imperfect balance. These are the \emph{equivalence classes} of measurement uncertainty, represented by the level curves of the surface.  The concentric contours on the bottom of \fig~\ref{fig:many_logons3d}B are the projections of some of the level curves.

\begin{figure*}[tbp]
\begin{minipage}[c]{.45\textwidth}
\centering
\includegraphics[width=.95\textwidth]{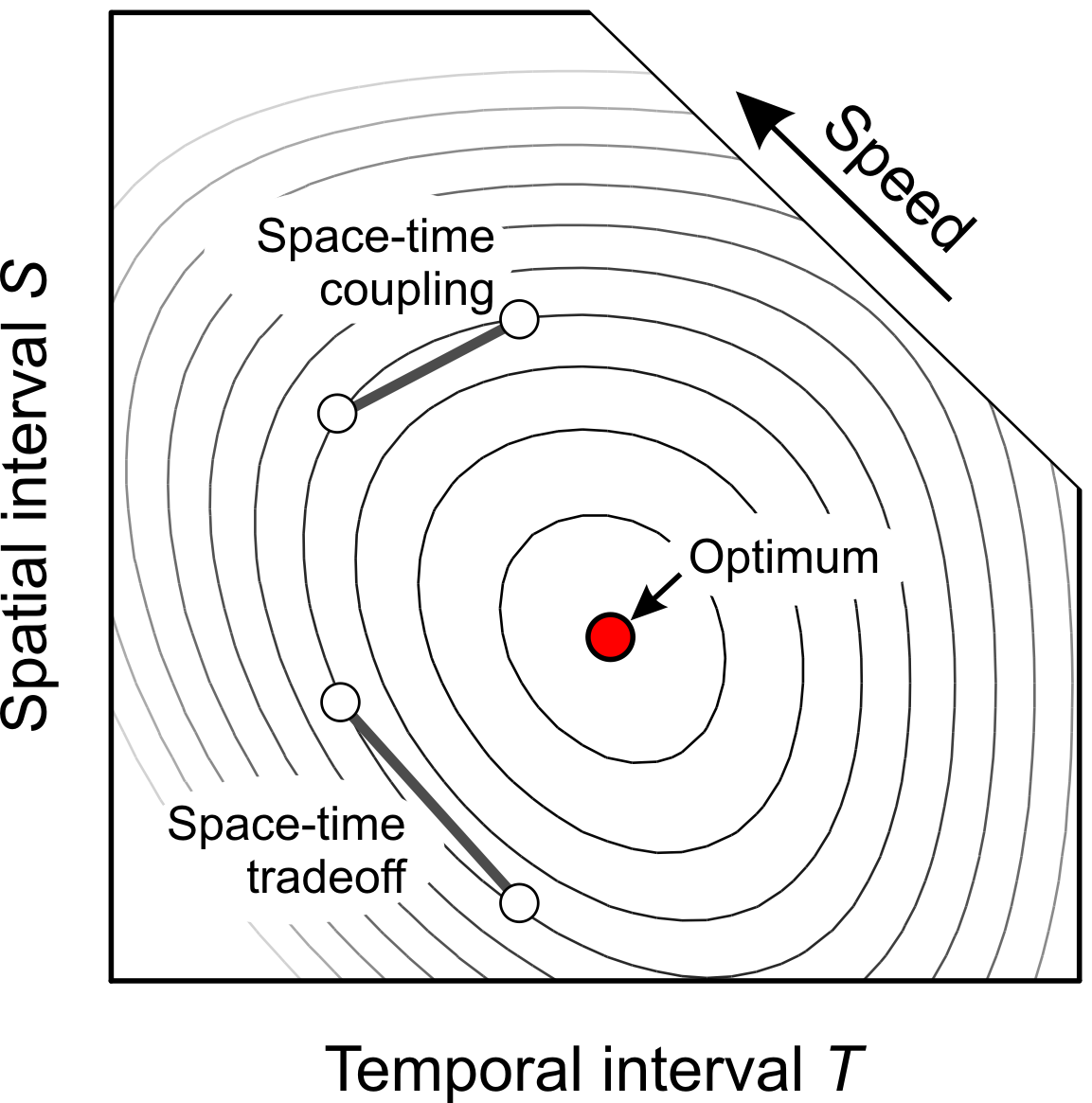}
\end{minipage}
\hfill
\begin{minipage}[c]{.47\textwidth}
\caption{\setstretch{1.0}
Equivalence classes of uncertainty. The contours represent equal measurement uncertainty (reproduced from the bottom panel of~\fig~\ref{fig:many_logons3d}B) and the red circle represents the minimum of uncertainty. The pairs of connected circles labeled ``space-time coupling" and ``space-time tradeoff" indicate why some studies of \am discovered different regimes of motion perception in different stimuli \cite{GepshteinKubovy2007,GepshteinTyukinKubovy2011}.}
\label{fig:eq_contours}
\end{minipage}
\end{figure*}

\subsection{Equivalence classes of uncertainty}

\noindent
Contours of equal measurement uncertainty are reproduced in \fig~\ref{fig:eq_contours} from the bottom of \fig~\ref{fig:many_logons3d}B. The pairs of connected circles indicate that the slopes of equivalence contours vary across the conditions of measurement. This fact has several interesting implications for the perception of visual motion.

First, if the equivalent conditions of motion perception were consistent with the equivalent conditions of uncertainty, then some lawful changes in the perception of motion would be expected for stimuli that activate sensors in different parts of the sensor space. This prediction was confirmed in studies of \am, which is the experience of motion from discontinuous displays, where the sequential views of the moving objects (the ``corresponding image parts") are separated by spatial ($\sigma$) and temporal ($\tau$) distances. Perceptual strength of \am in such displays was conserved: sometimes by changing $\sigma$ and $\tau$ in the same direction (both increasing or both decreasing), which is the regime of space-time coupling \cite{Korte1915}, and sometimes by trading off one distance for another: the regime of space-time tradeoff \cite{BurtSperling1981}. Gepshtein and Kubovy \cite{GepshteinKubovy2007} found that the two regimes of \am were special cases of a lawful pattern: one regime yielded to another as a function of speed, consistent with the predictions illustrated in ~\fig~\ref{fig:eq_contours}.

Second, the regime of space-time coupling undermines one of the cornerstones of the literature on visual perceptual organization: the \emph{proximity principle} of perceptual grouping \cite{Wertheimer1923,KubovyEtAl1998}. The principle is an experimental observations from the early days of the Gestalt movement, capturing the common observation that the strength of grouping between image parts depends on their distance: the shorter the distance the stronger the grouping. In space-time, the principle would hold if the strength of grouping had not changed, when increasing one distance ($\sigma$ or $\tau$) was accompanied by decreasing the other distance ($\tau$ or $\sigma$): the regime of tradeoff \cite{Koffka1963}. The fact that the strength of grouping is maintained by increasing both $\sigma$ and $\tau$, or by or decreasing both $\sigma$ and $\tau$,  is inconsistent with the proximity principle \cite{GepshteinTyukinKubovy2011}.

\subsection{Spatiotemporal interaction: speed}

\noindent
Now let us consider the interaction of the spatial and temporal dimensions of measurement. A key aspect of this interaction is the speed of stimulus variation: the rate of temporal change of stimulus intensity across spatial location. The dimension of speed has been playing a central role in the theoretical and empirical studies of visual perception \cite{Nakayama1985,WatsonAhumada1985,WeissEtAl2002}. Not only is the perception of speed crucial for the survival of mobile animals, but it also constitutes a rich source of auxiliary information for parsing the optical stimulation \cite{LonguetHiggins_Prazdny1981,LandyEtAl95}.

What is more, speed appears to play the role of  a control parameter in the organization of visual sensitivity. The shape of a large-scale characteristic of visual sensitivity (measured using continuous stimuli) is invariant with respect to speed \cite{Kelly1979,Kelly1994b}.  And a characteristic of the strength of perceived motion in discontinuous stimuli (giving rise to ``apparent motion") collapse onto a single function when plotted against speed \cite{GepshteinKubovy2007}.

From the present normative perspective, the considerations of speed measurement (combined with the foregoing considerations of measuring the location and frequency content) of visual stimuli have two pervasive consequences, which are reviewed in some detail next.  First, in a system optimized for the measurement of speed, the expected distribution of the quality of measurement has an invariant shape, distinct from the shape of such a distribution conceived before one has taken into account the measurement of speed (\fig~\ref{fig:eq_contours}). Second,  the \emph{dynamics} of visual measurement, and not only its static organization, will depend on the manner of interaction of the spatial and temporal aspects of measurement.

\begin{figure}[tbp]
\centering
\includegraphics[width=0.65\textwidth]{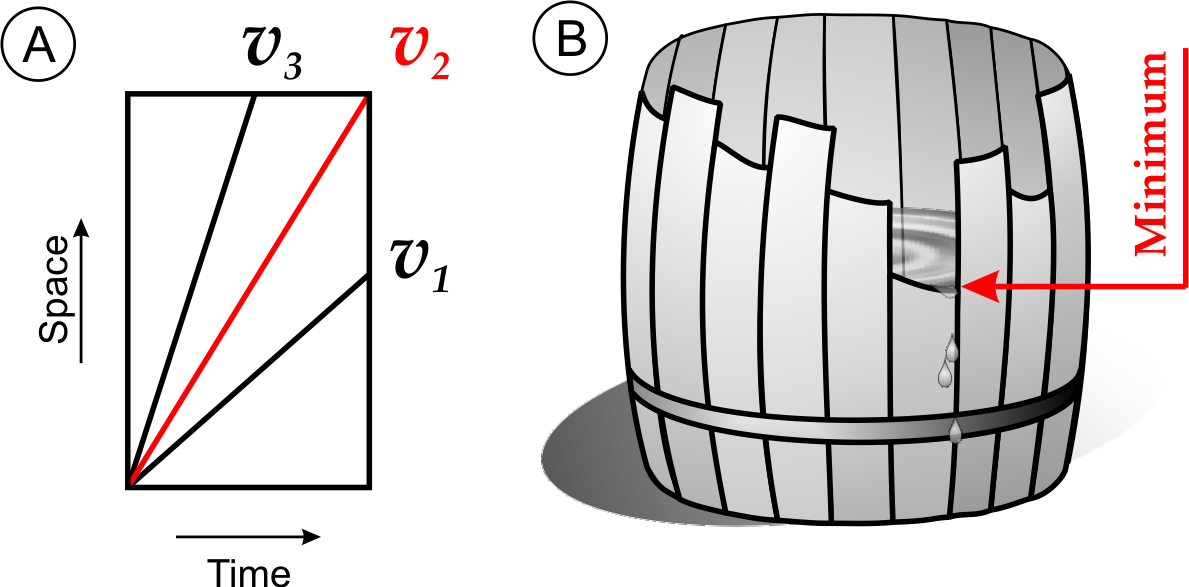}
\vspace{.1in}
\caption{\setstretch{1.0} Economic measurement of speed. (A)~The rectangle represents a sensor defined by spatial and temporal intervals~($S$ and~$T$). From considerations of parsimony, the sensor is more suitable for measurement of speed $v_2=S/T$ than~$v_1$ or~$v_3$ since no part of $S$ or~$T$ is wasted in measurement of $v_2$. (B)~Liebig's barrel. The shortest stave determines barrel's capacity. Parts of longer staves are wasted since they do not affect the capacity.}
\label{fig:barrel}
\end{figure}

In \figs~\ref{fig:many_logons3d}-\ref{fig:eq_contours}, a distribution of the expected uncertainty of  measurement was derived from a local constraint on measurement. The local constraint was defined separately for the spatial and temporal intervals of the sensor. The considerations of speed measurement add another constraint, which has to do with the relationship between the spatial and temporal intervals.

The ability to measure speed by a sensor defined by spatial and temporal intervals depends on the extent of these intervals. As it is shown in \fig~\ref{fig:barrel}A, different ratios of the spatial extent to the temporal extent make the sensor differently suitable for measuring different magnitudes of speed.

This argument is one consequence of the \emph{Law of The Minimum} \cite{Gorban2011:Bull}, illustrated in \fig~\ref{fig:barrel}B using Liebig's barrel. A broken barrel with the staves of different lengths can hold as much content as the shortest stave allows. Using the staves of different lengths is wasteful because a barrel with all staves as short as the shortest stave would do just as well.  In other words, the barrel's capacity is limited by the shortest stave.

Similarly, a sensor's capacity for measuring the speed is limited by the extent of its spatial and temporal intervals. The capacity is not used fully if the spatial and temporal projections of vector $v$ are larger or smaller than the spatial and temporal extents allow ($v_1$ and $v_3$ in \fig~\ref{fig:barrel}B). Just as the extra length of the long staves is wasted in the Liebig's barrel, the spatial extent of the sensor is wasted in measurement of $v_1$ and the temporal  extent  is wasted in measurement of $v_3$. Let us therefore start with the assumption that the sensor defined by the intervals $S$ and~$T$ is best suited for measuring speed $v=S/T$.

\section{Optimal conditions for motion measurement}
\label{sec:optimal_conditions}

\subsection{Minima of uncertainty}

\noindent
The optimal conditions of measurement are expected where the measurement uncertainty is the lowest.  Using a shorthand notation for the spatial and temporal partial derivatives of $U_{ST}$ in \eq~\ref{eq:u_j},
$\textstyle \pd U_{S} = \pd U_{ST} / \pd S$ and $\textstyle \pd U_T = \pd U_{ST} / \pd T$, the minimum of measurement uncertainty is the solution of
\begin{equation}
\textstyle
\pd U_T \thinspace d T + \pd U_S \thinspace d S = 0.
\label{eq:total_diff}
\end{equation}
The optimal condition for the entire space of sensors, disregarding individual speeds, is marked as the red point in \fig~\ref{fig:eq_contours}. To find the minima for specific speeds $v_i$, let us rewrite \eq~\ref{eq:total_diff} such that speed appears in the equation as an explicit term. By dividing each side of \eq~\ref{eq:total_diff} by $dT$, and using the fact that $v = dS/dT$ , it follows that
\begin{equation}
\textstyle
\pd U_S v_i + \pd U_T = 0.
\label{eq:eqlib1}
\end{equation}

The solution of \eq~\ref{eq:eqlib1} is a set of optimal conditions of measurement across speeds.  To illustrate the solution graphically,  consider the vector form of \eq~\ref{eq:eqlib1}, \ie the scalar product
\begin{equation}
\textstyle
\Big\langle\thinspace\mathbf{g}_{(T,S)}, \thickspace \mathbf{v}_{(T,S)}\thinspace\Big\rangle = 0,
\label{eq:eqlib_vectors}
\end{equation}
where the first term is the gradient of measurement uncertainty function,
\begin{equation}
\textstyle
\mathbf{g}_{(T,S)} = (\pd u_T,\pd u_S),
\label{eq:vector_g}
\end{equation}
and the second term is the speed,
\begin{equation}
\textstyle
\mathbf{v}_{(T,S)} = (T, vT),
\label{eq:vector_v}
\end{equation}
for sensors with parameters $(T, S)$. For now, assume that the speed to which a sensor is tuned is the ratio of spatial to temporal intervals ($v=S/T$) that define the logon of the sensor. (Normative considerations of speed tuning are reviewed in section \emph{Spatiotemporal interaction: speed}.)

\begin{figure*}[t!]
\centering
\includegraphics[width=1.00\textwidth]{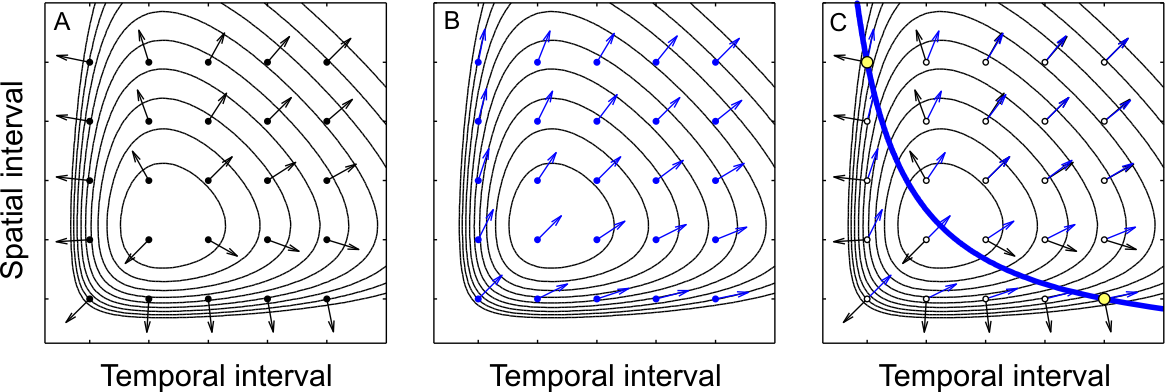}
\caption{\setstretch{1.0} Graphical solution of \eq~\ref{eq:eqlib_vectors} without integration of speed. (A)~Local gradients of measurement uncertainty $\mathbf{g}$. (B)~Speeds $\mathbf{v}$ to which the different sensors are tuned. (C)~Optimal conditions (blue curve) arise where $\mathbf{g}$ and $\mathbf{v}$ are orthogonal to one another (\eq~\ref{eq:eqlib_vectors}). The yellow circles are two examples of locations where the requirement of orthogonality is satisfied. (Arrow lengths are normalized to avoid clutter.)}
\label{fig:why_hyp_lin}
\vspace{.2in}
\includegraphics[width=1.00\textwidth]{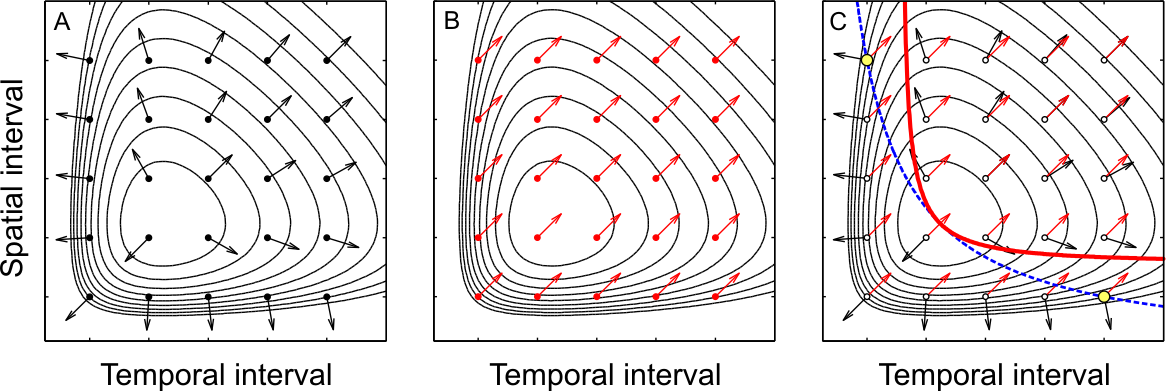}
\caption{\setstretch{1.0} Graphical solution of \eq~\ref{eq:eqlib_vectors} with integration of speed. (A)~Local gradient of measurement uncertainty $\mathbf{g}$ as in \fig~\ref{fig:why_hyp_lin}A. (B)~Speeds $\mathbf{v}$ integrated across multiple speeds. (C)~Now the optimal conditions (red curve) arise at locations different from those in \fig~\ref{fig:why_hyp_lin} (the blue curve is a copy from \fig~\ref{fig:why_hyp_lin}C).}
\label{fig:why_hyp_lin2}
\end{figure*}

The two terms of \eq~\ref{eq:eqlib_vectors} are shown in~\fig~\ref{fig:why_hyp_lin}: separately in panels~A-B and together in panel~C. The blue curve in panel~C represents the set of conditions where vectors $\mathbf{v}$ and $\mathbf{g}$  are orthogonal to one another, satisfying \eq~\ref{eq:eqlib_vectors}. This curve is the \emph{optimal set} for measuring speed while minimizing the uncertainty about signal location and content.

\begin{figure*}[tbp]
\centering
\includegraphics[width=1.00\textwidth]{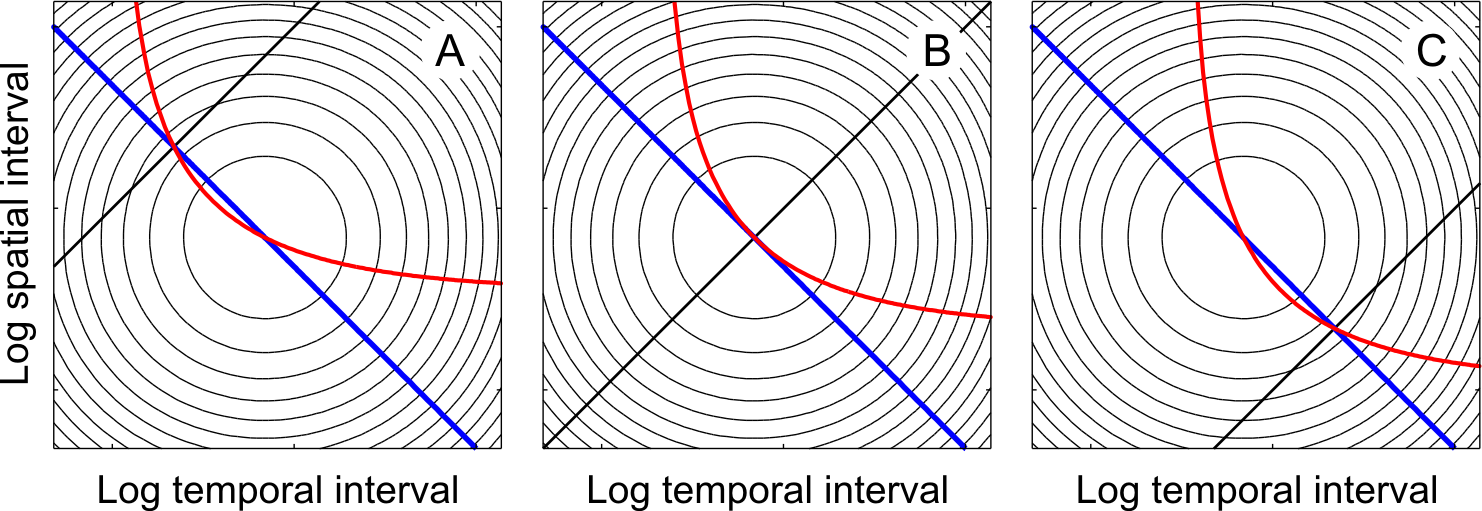}
\caption{\setstretch{1.0} Effect of expected stimulus speed.  The red and blue curves are the optimal sets derived in \figs~\ref{fig:why_hyp_lin}-\ref{fig:why_hyp_lin2}, now shown in logarithmic coordinates to emphasize that the ``integral" optimal set (red) has the invariant shape of a rectangular hyperbola, whereas the ``local" optimal set (blue) does not. From~A to~C, the expected stimulus speed (\eq~\ref{eq:expected_speed}) decreases, represented by the black lines. The position of the  integral optimal set changes accordingly.}
\label{fig:why_hyp_log}
\end{figure*}

\subsection{The shape of optimal set}

\noindent
The solution of \eq~\ref{eq:eqlib_vectors} was derived for speed defined at every point in the space of intervals (T,S): the blue arrows in \fig~\ref{fig:why_hyp_lin}B. This picture is an abstraction that disregards the fact that measurements are performed while the sensors integrate stimulation over sensor extent.  The solution of \eq~\ref{eq:eqlib_vectors} that takes this fact into account is described in \fig~\ref{fig:why_hyp_lin2}. The integration reduces differences between the directions of adjacent speed vectors (panel~B), and so the condition of orthogonality of $\mathbf{g}$ and $\mathbf{v}$ is satisfied at locations other than in \fig~\ref{fig:why_hyp_lin}.

The red curve \fig~\ref{fig:why_hyp_lin2}C is the \emph{integral optimal set} for measuring speed. This figure presents an extreme case, where speeds are integrated across the entire range of stimulation, as if  every sensor had access to the expected speed of stimulation across the entire range of stimulus speed:
\begin{equation}
v_{e} = \int_0^\infty p(v) \thinspace v \thinspace dv,
\label{eq:expected_speed}
\end{equation}
where $p(v)$ is the distribution of speed in the stimulation. At this extreme, every $\mathbf{v}$ is co-directional with the expected speed.

In comparison to the local optimal set (the blue curve in \fig~\ref{fig:why_hyp_lin2}C), many points of the integral optimal set (the red curve) are shifted away from the origin of the parameter space. The shift is small in the area of expected speed $v_e$ (the black line in \fig~\ref{fig:why_hyp_log}), yet the shift increases away from the expected speed, such that the integral optimal set has the shape of a hyperbola.

The position of the optimal set in the parameter space depends on the prevailing speed of stimulation \cite{GepshteinTyukinKubovy2007}, as \fig~\ref{fig:why_hyp_log} illustrates. This dependence is expected to be more pronounced in cases where the integration by receptive fields is large.

To summarize, the above argument has been concerned with how speed integration affects the optimal conditions for speed measurement. At one extreme, with no integration, the set of optimal conditions could have any shape. At the other extreme, with the scope of integration maximally large, the optimal set is a hyperbola. In between, the larger the scope of integration, the more the optimal set resembles a hyperbola. The position of this hyperbola in the parameter space depends on the prevalent speed of stimulation.

\begin{figure}[tbp]
\centering
\includegraphics[width=1.00\textwidth]{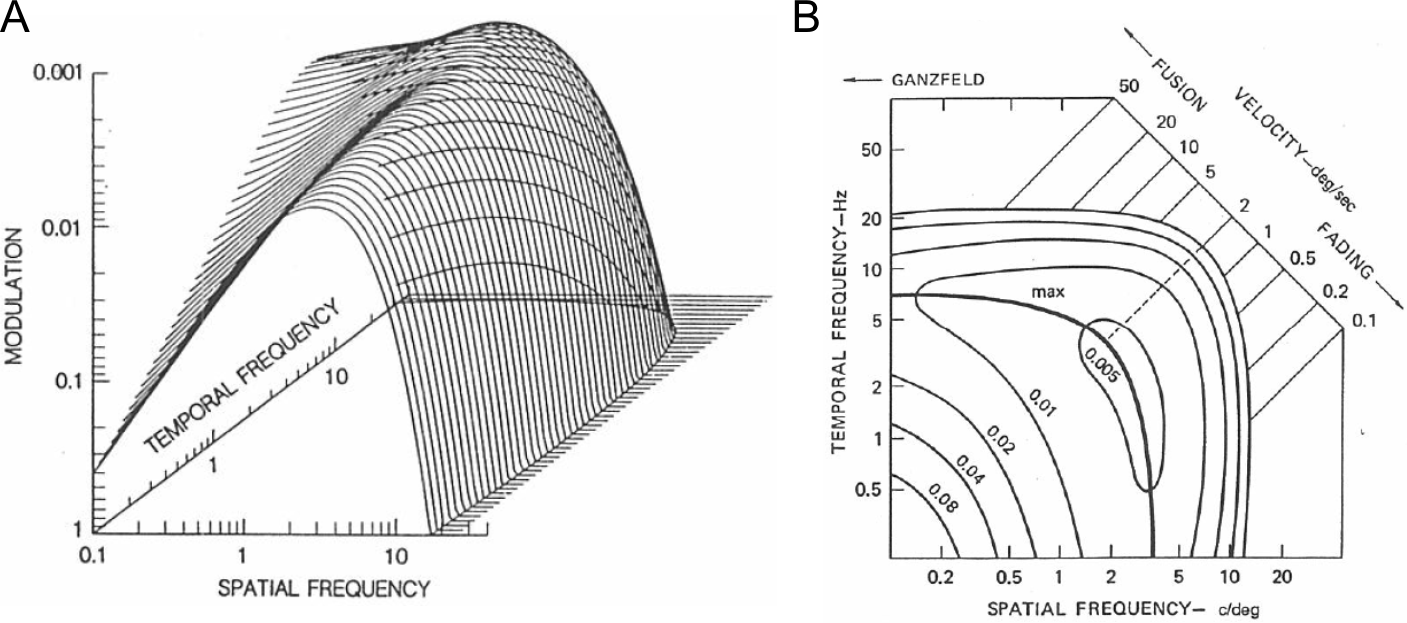}
\caption{\setstretch{1.0}
Human spatiotemporal contrast sensitivity function, shown as a surface in~A and a contour plot in~B. Conditions of maximal sensitivity across speeds form the thick curve labeled ``max." The maximal sensitivity set has the shape predicted by the normative theory: the red curve in \fig~\ref{fig:why_hyp_lin2}. The mapping from measurement intervals to stimulus frequencies is  explained in \cite{Nakayama1985,GepshteinTyukinKubovy2007}. Both panels are adopted from \cite{Kelly1979}.}
\label{fig:kelly}
\end{figure}

This argument has two significant implications. First, the distribution of resources in the visual system is predicted to have an invariant shape,  which is consistent with results of measurements in biological vision (\fig~\ref{fig:kelly}) using a variety of psychophysical tasks and stimuli  \cite{VanDoornKoenderink1982a,VanDoornKoenderink1982b,Nakayama1985,Laddis_sfn2011}.
Second, it implies that changes in statistics of stimulation will have a predictable effect on allocation of resources, helping the systems adapt to the variable stimulation, a theme developed in the next section.

\section{Sensor allocation}
\label{sec:allocation}

\subsection{Adaptive allocation}

\noindent Allocation of sensors is likely to depend on several factors that determine sensor usefulness, such as sensory tasks and properties of stimulation.  For example, when the organism needs to identify rather than localize the stimulus, large sensors are more useful than small ones. Allocation of sensors by their usefulness  is therefore expected to shift, for example as shown in \fig~\ref{fig:many_logons1d}C.

Such shifts of allocation are expected also because the environment is highly variable. To insure that sensors are not allocated to stimuli that are absent or useless, biological systems must monitor their environment and the needs of measurement. As the environment or needs change, the same stimuli become more or less useful.  The system must be able to reallocate its resources: change properties of  sensors such as to better measure useful stimuli.

Because of the large but limited pool of sensors at their disposal, real sensory systems occupy the middle ground between extremes of sensor ``wealth."  Such systems can afford some specialization but they cannot be wasteful. They are therefore subject to Gabor's uncertainty relation, but they can alleviate consequences of the uncertainty relation, selectively and to some extent, by allocating sensors to some important classes of stimuli. Allocation preferences of  such systems is expected to look like that in \fig~\ref{fig:many_logons1d}C, yet generalized to multiple stimulus dimensions.

To summarize, the above analysis suggests that sensory systems are shaped by constraints of measurement and the economic constraint of limited resources. This is  because the sensors of different sizes are ordered according to their usefulness in view of Gabor's uncertainty relation. These considerations are exceedingly simple in the one-dimensional analysis undertaken so far.  In a more complex case considered in the next section, this approach leads to nontrivial conclusions.  In particular, this approach helps to explain several puzzling phenomena in perception of motion and in motion adaptation.

\begin{figure*}[tb]
\centering
\includegraphics[width=\textwidth]{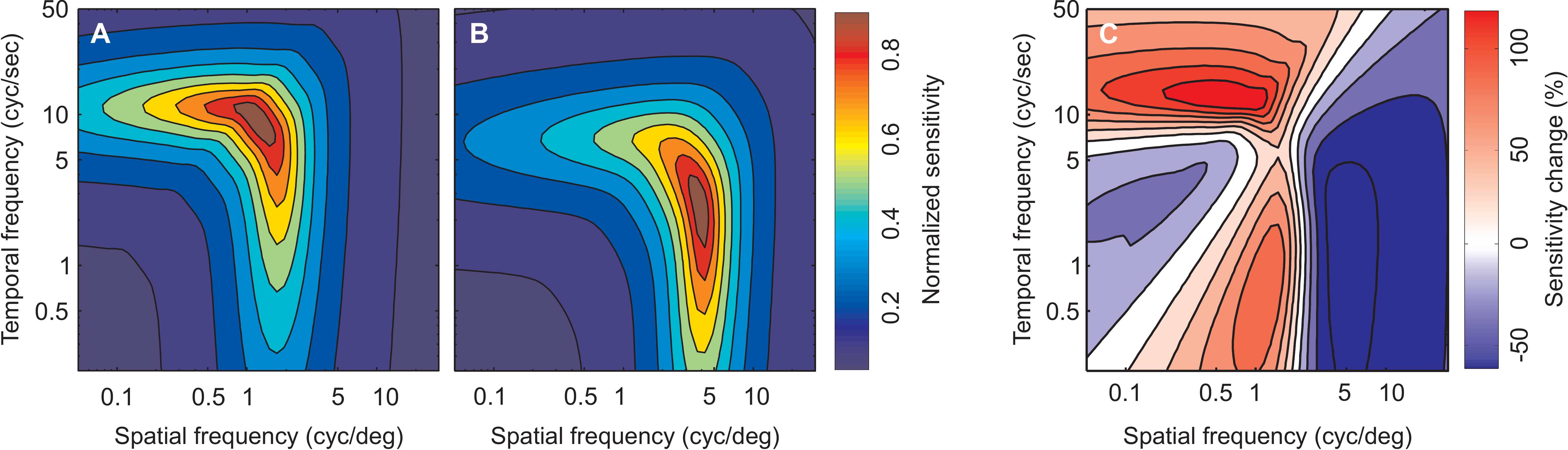}
\caption{\setstretch{1.0}
Predictions for adaptive reallocation of sensors. (A--B)~Sensitivity maps predicted for two stimulus contexts: dominated by high speed in~A and low speed in~B. The color stands for normalized sensitivity. (C)~Sensitivity changes computed as $100 \times a/b$ where $a$ and $b$ are map entries in~A and~B, respectively.  Here, the color stands for sensitivity change: gain in red and loss in blue.}
\label{fig:adaptation2}
\end{figure*}
\nocite{Nakayama1985}

A  prescription has been derived for how \rfs of different spatial and temporal extents ought to be distributed across the full range of visual stimuli. By this prescription, changes in usefulness of stimuli are expected to cause changes in \rf allocation. Now consider some specific predictions of how the reallocation of resources is expected to bring about systematic changes in spatiotemporal visual sensitivity. Because the overall amount of resources in the system is limited, an improvement of visual performance  (such as a higher sensitivity) at some conditions will be accompanied  by a deterioration of performance (a lower sensitivity) at other conditions, leading to counterintuitive patterns of sensitivity change.

Assuming that equivalent amounts of resources should be allocated to equally useful stimuli, when certain speeds become more prevalent or more important for perception than other speeds, the visual system is expected to allocate more resources to the more important speeds.

For example, \fig~\ref{fig:adaptation2}A--B contains maps of spatiotemporal sensitivity computed for two environments, with high and low prevailing speeds. \fig~\ref{fig:adaptation2}C is a summary of differences between the sensitivity maps: The predicted changes form well-defined foci of increased performance and large areas of decreased performance. Gepshtein et al.~\cite{Gepshtein_etal:2013} used intensive psychometric methods \cite{LesmesEtAl_vss2009} to measure the entire spatiotemporal contrast sensitivity function  in different statistical ``contexts" of stimulation. They found that sensitivity changes were consistent with the predictions illustrated in  \fig~\ref{fig:adaptation2}.

These results suggest a simple resolution to some long-standing puzzles in the literature on motion adaptation. In early theories, adaptation was viewed as a manifestation of neural fatigue. Later theories were more pragmatic, assuming that sensory adaptation is the organism's attempt to adjust to the changing environment \cite{SakittBarlow1982,Laughlin1989,Wainwright1999,LaughlinSejnowski2003}. But evidence supporting this view has been scarce and inconsistent. For example, some studies showed that perceptual performance improved at the adapting conditions, but other studies reported the opposite \cite{CliffordWenderoth1999,krekelberg_et_al2006}. Even more surprising were systematic changes of performance for stimuli very different from the adapting ones  \cite{krekelberg_et_al2006}.  According to the present analysis, such local gains and losses of sensitivity are expected in a visual system that seeks to allocate its limited resources in face of uncertain and variable stimulation (\fig~\ref{fig:adaptation2}).  Indeed, the pattern of  gains and losses of sensitivity manifests an optimal adaptive visual behavior.

This example illustrates that in a system with scarce resources optimization of performance will lead to reduction of sensitivity to some stimuli. This phenomenon is not unique to sensory adaptation \cite{Gepshtein2009}. For example,  demanding tasks may cause impairment of visual performance for some stimuli, as a consequence of  task-driven reallocation of visual resources \cite{YeshurunCarrasco1998,YeshurunCarrasco2000}.

\subsection{Mechanism of adaptive allocation}

From the above it follows that the shape of spatiotemporal sensitivity function, and also transformations of this function, can be understood by studying the uncertainties implicit to visual measurement. This idea received further support from simulations of a visual system equipped with thousands of independent (uncoupled) sensors, each having a spatiotemporal \rf \cite{JuricaEtAl2007,JuricaEtAl2013}.

In these studies, spatiotemporal signals were sampled from known statistical distributions.  \Rfs parameters were first distributed at random. They were then updated according to a generic rule of synaptic plasticity \cite{Hebb:1949,Bienenstock:1982,PaulsenSejnowski:2000,Bi:2001}. Changes of \rf amounted to  small random steps  in the parameters space, modeled as stochastic fluctuations of  the spatial and temporal extents of \rfs.  Step length was proportional to the (local) uncertainty of measurement by individual \rfs. The steps were small where the uncertainty was low, and \rfs changed little.  Where the uncertainty was high, the steps were larger, so the \rfs tended to escape the high-uncertainty regions. The stochastic behavior led to a ``drift" of \rfs in the direction of low uncertainty of measurement \cite{JuricaEtAl2013}, predicted by standard stochastic methods \cite{Gardiner1996}, as if the system sought stimuli that could be measured reliably (\emph{cf}.~\cite{VergassolaEtAl2007}).

Remarkably, the independent stochastic changes of \rfs (their uncoupled ``stochastic tuning") steered the system toward the distribution  of \rf parameters predicted by the normative theory, and forming the distribution observed in human vision (\fig~\ref{fig:kelly}).  When the distribution of stimuli changed,  mimicking a change of sensory environment, the system was able to spontaneously discover an  arrangement of sensors optimal for the new  environment, in agreement with the predictions illustrated in \fig~\ref{fig:adaptation2} \cite{Gepshtein_sfn2010}. This is an example of how efficient allocation of resources can emerge in sensory systems by way of self-organization, enabling a highly adaptive sensory behavior in face of the variable (and sometimes unpredictable) environment.

\subsection{Conclusions}

A study of allocation of limited resources for motion sensing across multiple spatial and temporal scales revealed that the optimal allocation entails a shape of the distribution of sensitivity similar to that found in human visual perception. The similarity suggested that several previously puzzling phenomena of visual sensitivity, adaptation, and perceptual organization have simple principled explanations. Experimental studies of human vision have confirmed the predictions for sensory adaptation. Since the optimal allocation is readily implemented in self-organizing neural networks by means of unsupervised leaning and stochastic optimization, the present approach offers a framework for neuromorphic design of multiscale sensory systems capable of automated efficient tuning to the varying optical environment.

\vspace{.25in}
\noindent {\bf Acknowledgments}

\vspace{.1in}
\noindent This work was supported by the European Regional Development Fund,  National Institutes of Health
Grant~EY018613, and Office of Naval Research Multidisciplinary University Initiative Grant~N00014-10-1-0072.

\afterpage{\rhead{\it REFERENCES}}
\afterpage{\lhead{ }}
\newpage

\bibliographystyle{splncs}

\afterpage{\rhead{\it APPENDICES}}
\afterpage{\lhead{ }}
\newpage
\section{Appendices}
\renewcommand{\thefigure}{A1.\arabic{figure}}
\setcounter{figure}{0}
\renewcommand{\theequation}{A1.\arabic{equation}}
\setcounter{equation}{0}
\renewcommand{\thefigure}{A.\arabic{figure}}
\setcounter{figure}{0}
\renewcommand{\theequation}{A2.\arabic{equation}}
\renewcommand{\theequation}{A.\arabic{equation}}
\setcounter{equation}{0}
\subsection{Appendix 1. Additivity of uncertainty}

For the sake of simplicity, the following derivations concern the stimuli that can be modeled by integrable functions $I:\Real\rightarrow\Real$ of one variable $x$. Generalizations to functions of more than one variable are straightforward. Consider two quantities:
\begin{itemize}

\item Stimulus location on $x$, where $x$ can be  space or time, the ``location" indicating  respectively ``where" or ``when" the stimulus occurred.
\vspace{.1in}
\item Stimulus content on $f_x$, where $f_x$ can be spatial or temporal frequency of stimulus modulation.
\end{itemize}
Suppose a sensory system is equipped with many measuring devices (``sensors"), each used to estimate both stimulus location and frequency content from ``image" (or ``input") $I(x)$. Assume that the outcome of measurement is a random variable with probability density function $p(x,f)$.

Let
\begin{equation}\label{sec:means}
\begin{split}
p_x(x) & =\int p(x,f)df, \\
p_f(f)  & = \int p(x,f)dx
\end{split}
\end{equation}
be the (marginal) means of $p(x,f)$ on dimensions $x$ and $f_x$ (abbreviated as $f$).

It is sometimes assumed that sensory systems ``know" $p(x,f)$, which is not true in general. Generally, one can only know (or guess) some properties of $p(x,f)$, such as its mean and variance. Reducing the chance of gross error due to the incomplete information about $p(x,f)$ is accomplished by a conservative strategy: finding the \emph{minima} on the function of \emph{maximal} uncertainty, \ie using a minimax approach \cite{vonNeumann1928,LuceRaiffa57}.

The minimax approach is implemented in two steps. The first step is to find such $p_x(x)$ and $p_f(f)$ for which measurement uncertainty is maximal. (Uncertainty is characterized conservatively, in terms of variance alone \cite{Gabor1946}.) The second step is to find the condition(s) at which the function of maximal uncertainty has the smallest value: the minimax point.

Maximal uncertainty is evaluated using the well-established definition of entropy \cite{Shannon1948} (cf.~\cite{Jaynes1957,Gorban2013:CMA}):
\begin{equation*}\label{eq:entropy:1}
H(X,F)=- \int p(x,f)\log p(x,f) dx \, df.
\end{equation*}
According to the \emph{independence bound on entropy} (Theorem~2.6.6 in \cite{CoverThomas2006}):
\begin{equation}\label{eq:entropy:2}
H(X,F)\leq H(X)+H(F)=H^{\ast}(X,F),
\end{equation}
where
\begin{equation*}\label{eq:entropy:3}
\begin{split}
H(X) =  & - \int p_x(x)\log p_x(x) dx,  \\
H(F) =  & - \int p_f(f)\log p_f(f) df.
\end{split}
\end{equation*}
Therefore, the uncertainty of measurement cannot exceed
\begin{equation}\label{eq:entropy:worst}
\begin{split}
H^{\ast}(X,F)  =  & - \int p_x(x)\log p_x(x)dx \\
                              & - \int p_f(f)\log p_f(f) df.
\end{split}
\end{equation}
\eq~\ref{eq:entropy:worst} is the ``envelope" of maximal measurement uncertainty: a ``worst-case" estimate.

By the Boltzmann theorem on maximum-entropy probability distributions \cite{CoverThomas2006}, the maximal entropy of probability densities with fixed means and variances is attained when the functions are Gaussian. Then, maximal entropy is a sum of their variances \cite{CoverThomas2006} and
\begin{equation*}\label{eq:entropy:4}
\begin{split}
p_x(x) & =\frac{1}{\sigma_x \sqrt{2\pi}} e^{-x^2/2\sigma_x^2},  \\
p_f(f)  & =\frac{1}{\sigma_f \sqrt{2\pi}} e^{-f^2/2\sigma_f^2},
\end{split}
\end{equation*}
where $\sigma_x$ and $\sigma_f$ are the standard deviations. Then maximal entropy is
\begin{equation} \label{eq:sum:variances}
H = \sigma_x^2 + \sigma_f^2.
\end{equation}
That is, when $p(x,f)$ is unknown, and all one knows about marginal distributions $p_x(x)$ and  $p_f(f)$ is their means and variances, the maximal uncertainty of measurement is the sum of variances of the estimates of $x$ and $f$. The following {\it minimax} step is to find the conditions of measurement at which the sum of variances is the smallest.

\subsection{Appendix 2. Improving resolution by multiple sampling}

How does an increased allocation of resources to a specific condition of measurement improve resolution (spatial or temporal) at that condition? Consider set $\Psi$ of sampling functions
\[\psi(s\sigma+\delta), \thickspace \sigma\in\Real, \thickspace
\sigma>0, \thickspace \delta\in\Real,\] where $\sigma$ is a
scaling parameter and $\delta$ is a translation parameter. For a broad class of functions $\psi(\cdot)$, any element of $\Psi$ can be obtained by addition of weighted and shifted $\psi(s)$. The following argument proves that {\it any} function from a sufficiently broad class that includes $\psi(s\sigma+\delta)$ can be represented by a weighted sum of translated replicas of $\psi(s)$.

Let $\psi^{\ast}(s)$ be a continuous function that can be expressed as a sum of a converging series of harmonic functions:
\[
\psi^{\ast}(s)=\sum_{i} a_i \cos(\omega_i s) + b_i \sin(\omega_i s).
\]
For example, Gaussian sampling functions of arbitrary widths can be expressed as a sum of $\cos(\cdot)$ and $\sin(\cdot)$. Let us show that, if $|\psi(s)|$ is Riemann-integrable, \ie if
\[
-\infty<\int_{-\infty}^{\infty}|\psi(s)|ds <\infty,
\]
and its Fourier transform, $\widehat{\psi}$, does not vanish for all $\omega\in\Real$: $\widehat{\psi}(\omega)\neq 0$ (\ie its spectrum has no ``holes"), then the following expansion of $\psi^\ast$ is possible
\begin{equation}\label{eq:resolution:0}
\psi^{\ast}(s)=\sum_{i} c_i \psi(s+d_i) + \varepsilon(s),
\end{equation}
where $\varepsilon(s)$ is a residual that can be arbitrarily small.  This goal is attained by proving identities
\begin{equation}\label{eq:expansion:sampling}
\begin{split}
\cos(\omega_0 s)&=\sum_{i} c_{i,1} \psi(s+d_{i,1}) + \varepsilon_1(s),\\
\sin(\omega_0 s)&=\sum_{i} c_{i,2} \psi(s+d_{i,2}) + \varepsilon_2(s),
\end{split}
\end{equation}
where $c_{i,1}$, $c_{i,2}$ and $d_{i,1}$, $d_{i,2}$ are real numbers, while $\varepsilon_1(s)$ and $\varepsilon_2(s)$ are arbitrarily small residuals.

First, write the Fourier transform of $\psi(s)$ as
\[
\widehat{\psi}(\omega)=\int_{-\infty}^{\infty} \psi(s)e^{-i\omega s}ds,
\]
and multiply both sides of the above expression by $e^{i\omega_0\upsilon}$:
\begin{equation}\label{eq:resolution:1}
e^{i \omega_0\upsilon}\widehat{\psi}(\omega)=e^{i \omega_0\upsilon}\int_{-\infty}^{\infty} \psi(s)e^{-i\omega s}ds=\int_{-\infty}^{\infty} \psi(s)e^{-i(\omega s - \omega_0\upsilon)}ds.
\end{equation}
Change the integration variable:
\[
x=\omega s - \omega_0\upsilon \Rightarrow dx = \omega ds, \ s=\frac{x+\omega_0\upsilon}{\omega},
\]
such that \eq~\ref{eq:resolution:1} transforms into
\[
e^{i \omega_0\upsilon}\widehat{\psi}(\omega)=\frac{1}{\omega}\int_{-\infty}^{\infty} \psi\left(\frac{x+\omega_0\upsilon}{\omega}\right)e^{-i x}dx.
\]
Notice that $\widehat{\psi}(\omega)=a(\omega)+i b(\omega)$. Hence
{\small
\[
e^{i \omega_0 \upsilon}\widehat{\psi}(\omega)=e^{i \omega_0 \upsilon}(a(\omega)+i b(\omega))=(\cos(\omega_0 \upsilon)+ i \sin(\omega_0 \upsilon))(a(\omega)+i b(\omega))
\]}
and
\[
\begin{split}
e^{i \omega_0 \upsilon}\widehat{\psi}(\omega)&=(\cos(\omega_0\upsilon)a(\omega)-\sin(\omega_0\upsilon)b(\omega)) + i (\cos(\omega_0\upsilon) b(\omega)+\sin(\omega_0\upsilon)a(\omega)).
\end{split}
\]
Since $\widehat{\psi}(\omega)\neq 0$ is assumed for all $\omega\in\Real$, then $a(\omega)+i b(\omega)\neq 0$. In other words, either $a(\omega)\neq 0$ or $b(\omega)\neq 0$ should hold. For example, suppose that $a(\omega)\neq 0$. Then
\[
Re\left(e^{i \omega_0 \upsilon}\widehat{\psi}(\omega)\right)+\frac{b(\omega)}{a(\omega)}Im\left(e^{i \omega_0 \upsilon}\widehat{\psi}(\omega)\right)= \cos(\omega_0\upsilon)\left(\frac{a^2(\omega)+b^2(\omega)}{a(\omega)}\right).
\]
Therefore
\begin{equation}\label{eq:resolution:2}
\begin{split}
\cos(\omega_0 \upsilon)&=\left(\frac{a(\omega)}{a^2(\omega)+b^2(\omega)}\right) Re \left(\frac{1}{\omega}\int_{-\infty}^{\infty} \psi\left(\frac{x+\omega_0\upsilon}{\omega}\right)e^{-i x}dx\right)\\
 &+ \left(\frac{b(\omega)}{a^2(\omega)+b^2(\omega)}\right)Im\left(\frac{1}{\omega}\int_{-\infty}^{\infty} \psi\left(\frac{x+\omega_0\upsilon}{\omega}\right)e^{-i x}dx\right).
 \end{split}
\end{equation}
Because function $\psi(s)$ is Riemann-integrable, the integrals in \eq~\ref{eq:resolution:2} can be approximated as
\begin{equation}\label{eq:resolution:3}
Re \left(\frac{1}{\omega}\int_{-\infty}^{\infty} \psi\left(\frac{x+\omega_0\upsilon}{\omega}\right)e^{-i x}dx\right)= \frac{\Delta}{\omega} \sum_{k=1}^N \psi\left(\frac{x_k+\omega_0\upsilon}{\omega}\right)\cos(x_k) + \frac{\bar{\varepsilon}_1(\upsilon,N)}{2\omega},
\end{equation}
\begin{equation}\label{eq:resolution:4}
Im \left(\frac{1}{\omega}\int_{-\infty}^{\infty} \psi\left(\frac{x+\omega_0\upsilon}{\omega}\right)e^{-i x}dx\right)= \frac{\Delta}{\omega} \sum_{p=1}^N \psi\left(\frac{x_p+\omega_0\upsilon}{\omega}\right)\sin(x_p) + \frac{\bar{\varepsilon}_1(\upsilon,N)}{2\omega},
\end{equation}
where $\ x_k$ and $x_p$ are some elements of $\Real$. To complete the proof, denote
\[
\begin{split}
c_{2i,1}&=\frac{\Delta}{\omega} \frac{a(\omega)}{a^2(\omega)+b^2(\omega)}\cos(x_i), \quad
c_{2i-1,1}=\frac{\Delta}{\omega} \frac{a(\omega)}{a^2(\omega)+b^2(\omega)}\sin(x_i),\\
d_{2i,1}&=d_{2i-1,1}=\frac{x_i}{\omega}.
\end{split}
\]
From
\eqs~\ref{eq:resolution:2}--\ref{eq:resolution:4} it follows that
\[
\cos(\omega_0\upsilon)=\sum_{j=1}^{2N} c_{j,1} \psi\left(\frac{\omega_0\upsilon}{\omega}+d_{j,1}\right) + \varepsilon_1(\upsilon,N).
\]
Given that $\widehat{\psi}(\omega)\neq 0$ for all $\omega$, and letting $\omega=\omega_0$, it follows that
\begin{equation}\label{eq:resolution:5}
\cos(\omega_0\upsilon)=\sum_{j=1}^{2N} c_{j,1} \psi\left(\upsilon+d_{j,1}\right) + \varepsilon_1(\upsilon,N),
\end{equation}
where
\begin{equation}\label{eq:resolution:6}
\varepsilon_1(\upsilon,N)=\bar{\varepsilon}_1(\upsilon,N)\frac{a(\omega_0)}{2\omega_0 (a^2(\omega_0)+b^2(\omega_0))}.
\end{equation}
An analogue of  \eq~\ref{eq:resolution:5} for
$\sin(\omega_0\upsilon)$ follows from
$\sin(\omega_0\upsilon)=\cos(\omega_0\upsilon+\pi/2)$. This
completes the proof of \eq~\ref{eq:expansion:sampling} and
hence of \eq~\ref{eq:resolution:0}.

\end{document}